\documentclass[a4paper]{llncs}

\usepackage{alltt}
\usepackage{caption}
\usepackage{paralist}
\usepackage{nicefrac}
\usepackage{booktabs}
\usepackage{algorithm}
\usepackage{algpseudocode}
\usepackage{xspace}
\usepackage{amsmath}
\usepackage{marvosym}
\usepackage{wasysym}
\usepackage{verbatim}
\usepackage{subfigure}
\usepackage{hyperref}
\usepackage{multirow}
\usepackage{cite}
\usepackage[capitalize]{cleveref}
\usepackage[per-mode=symbol,detect-all]{siunitx}
\usepackage{graphics}
\usepackage{amssymb}
\usepackage{wrapfig}
\usepackage{enumitem}
\usepackage{xcolor}
\usepackage{booktabs}
\usepackage{cancel}
\usepackage{tabularx}

\definecolor{navy}{RGB}{0,0,128}

\newcommand{\relu}{\text{ReLU}\xspace{}}
\newcommand{\cost}{\text{cost}\xspace{}}
\newcommand{\bestCost}{\text{bestCost}\xspace{}}
\newcommand{\bc}{\text{bestChange}\xspace{}}

\newcommand\hcancel[2][black]{\setbox0=\hbox{$#2$}%
	\rlap{\raisebox{.45\ht0}{\textcolor{#1}{\rule{\wd0}{1pt}}}}#2} 

\usepackage{eqparbox}

\newcommand{\tool}{\textsc{3M-DNN}}

\usepackage{tikz,ifthen,pgfplots}
\usetikzlibrary{arrows,trees,backgrounds,automata,shapes,decorations,plotmarks,fit,calc,positioning,shadows,chains,decorations}
\tikzstyle{every pin edge}=[<-,shorten <=1pt]
\tikzstyle{neuron}=[circle,fill=black!25,minimum size=17pt,inner sep=0pt]
\tikzstyle{input neuron}=[neuron, fill=green!50]
\tikzstyle{output neuron}=[neuron, fill=red!50]
\tikzstyle{hidden neuron}=[neuron, fill=blue!50]
\tikzstyle{sep neuron}=[neuron, fill=orange!50]
\tikzstyle{annot} = [text width=6em, text centered]
\tikzstyle{nnedge} = [-{stealth},shorten >=0.1cm, shorten <=0.05cm,line width=0.8pt,black]
\usetikzlibrary{calc}

\newcommand{\norm}[1]{\left\lVert#1\right\rVert}

\begin{document}
	\title{Minimal Multi-Layer Modifications of \\
		Deep Neural Networks}

	\author{
		Idan Refaeli \and
		Guy Katz
	}
	\institute{
		The Hebrew University of Jerusalem, Jerusalem, Israel \\
		\email{ \{idan.refaeli, g.katz\}@mail.huji.ac.il}\\
	}
	
	\maketitle
	
	\begin{abstract}
		Deep neural networks (DNNs) have become increasingly popular
		in recent years. However, despite their many successes, DNNs
		may also err and produce incorrect and
		potentially fatal outputs in safety-critical settings, such
		as autonomous driving, medical diagnosis, and airborne
		collision avoidance systems. Much work has been put into
		detecting such erroneous behavior in DNNs, e.g., via testing
		or verification, but removing these errors after their
		detection has received lesser attention. We present here a
		new tool, called \tool{}, for \emph{repairing} a given DNN,
		which is known to err on some set of inputs. The novel
		repair procedure implemented in \tool{} computes a
		modification to the network's weights that corrects its
		behavior, and attempts to minimize this change via a
		sequence of calls to a backend, black-box DNN verification
		engine. To the best of our knowledge, our method is the
		first one that allows repairing the network by
		simultaneously modifying multiple layers. This is achieved
		by splitting the network into sub-networks, and applying a
		single-layer repairing technique to each component. We
		evaluated \tool{} tool on an extensive set of benchmarks,
		obtaining promising results.
	\end{abstract}
	
	\section{Introduction}

	The popularity of \emph{deep neural networks}
	(\emph{DNNs})~\cite{GoBeCoBe16} has increased significantly
	over the past few years. DNNs are machine-learned artifacts,
	trained using a finite training set of examples; and they are
	capable of correctly handling previously-unseen inputs. DNNs
	have shown great success in many application domains, such as
	image recognition~\cite{KrSuHi12, CiMeSc12}, audio
	transcription~\cite{SeLiYu12}, language
	translation~\cite{SiKuDaSiRaJa17}, and even in safety-critical
	domains such as medical
	diagnosis~\cite{KeGoCaVaLiBaMcYaWuYa18}, autonomous
	driving~\cite{BoDeDwFiFlGoJaMoMuZhZhZhZi16}, and airborne
	collision avoidance~\cite{JuLoBrOwKo16}.
	
	Despite their evident success, DNNs can sometimes contain
	bugs. This has been demonstrated repeatedly: in one famous
	example, Goodfellow et al.~\cite{GoShSz14} showed that slight
	perturbations to a DNN's input could lead to misclassification
	--- a phenomenon now known as susceptibility to
	\emph{adversarial perturbations}. In another case, Liu et
	al.~\cite{LiMaAaLeZhWaZh17} showed how DNNs are vulnerable to
	Trojan attacks. These issues, and others, combined with the
	increasing integration of DNNs into safety-critical systems,
	have created a surge of interest in establishing their
	correctness. A great deal of effort has been put into
	developing methods for testing DNNs~\cite{SuHuKrShHiAs18},
	and, more recently, also into verifying
	them~\cite{KaHuIbJuLaLiShThWuZeDiKoBa19, WaPeWhYaJa18, Ehlers2017}.
	These verification methods could play a significant role in
	the future certification of DNN-based systems.
	
	Here, we deal with the case where we already know that a given
	DNN is malfunctioning; specifically, we assume we have a
	finite set of concrete inputs which are handled erroneously
	(discovered by testing, verification, or any other method).
	In this situation, we would like to \emph{modify} the network,
	so that it produces correct predictions for these inputs. A
	na\"ive approach for accomplishing this is to add these faulty
	inputs to the training set used to create the DNN, and then
	retrain it, but this is often too computationally
	expensive~\cite{Ha19}. Also, retraining may change the network
	significantly, potentially introducing new bugs on inputs that
	were previously correctly handled. Finally, retraining might be
	impossible when the original training set is inaccessible,
	e.g., due to its privacy or sensitivity~\cite{JuLoBrOwKo16}.
	
	Instead, we advocate an approach that requires no retraining,
	and which has recently gained some
	attention~\cite{DoSuWaWaDa20, UsGoSuNoPa21, LeFrMaFrPaPa21,
		YaYaTrHoJoPo21}: we present a new tool, called \tool{}
	(\emph{M}inimal, \emph{M}ulti-layer \emph{M}odifications
	for \emph{DNNs}), which can directly find a modification to
	the network and correct the erroneous behavior. In this
	context, a modification means changing the networks
	\emph{weights} --- the set of real values that determine the
	DNN's output, and which are initially selected during
	training. Further, because we assume the original network is
	mostly correct, we seek to find a modification which is also
	\emph{minimal}. The motivation is that such a change would
	maintain as much as possible of the network's behavior on
	other inputs. In other words, our goal is to improve the DNN's
	overall \emph{accuracy} --- the percentage of correctly
	handled inputs, which is normally measured with respect to a
	\emph{test set} of examples --- by improving its handling of
	problematic inputs, and without harming its handling of other
	inputs.
	
	A DNN is, by definition, a \emph{layered} artifact; and to the
	best of our knowledge, all previous work on finding
	minimal modifications to a DNN's weights focused on changing
	the weights of a single layer~\cite{GoAdKeKa20, DoSuWaWaDa20,
		UsGoSuNoPa21}. Intuitively, and as we later demonstrate,
	this significant restriction could prevent one from finding
	potentially smaller (and thus preferable) changes to the
	network. In \tool{}, we seek to lift this restriction by proposing
	and implementing a novel method for the \emph{multi-layer} modification of a
	DNN, with the goal of finding smaller modifications than could
	be otherwise possible. The key idea of our approach is to
	split the network into multiple sub-networks along certain
	layers, which we refer to as \emph{separation layers}; and
	then attempt to find a minimal change for each of these
	sub-networks separately, in a way that brings about the
	desired overall change to the network.
	
	More concretely, \tool{} is comprised of two logical
	levels. In the top, \emph{search level}, the tool conducts a heuristic
	search through possible changes to the values computed by the
	separation layers. Each possible change to these values that
	we consider, translates into a possible fix to the DNN; it
	naturally gives rise to a sequence of problems on the bottom,
	\emph{single-layer modification level}, each involving a
	single sub-network. Solving these single-layer modification
	problems can be performed using existing techniques; and the
	changes discovered to the sub-networks modify the values of
	the separation layers as selected by the top level. Thus, the
	process as a whole allows \tool{} to reduce the problem of
	multi-layer changes into a sequence of single-layer change
	problems, which can be dispatched using existing DNN
	modification tools as backends.
	
	In its search for a minimal change, \tool{} alternates between the
	two levels: each time the top-level examines a potential
	change to the separation layers, and invokes the lower level
	in order to compute the overall cost of using that change (by
	combining the costs of changing each individual
	sub-network). The top-level always maintains the minimal
	change it has encountered so far, and uses search heuristics
	in order to find new, better options. The search space is
	infinite, and so our tool is \emph{anytime} --- it is
	designed to be run with a timeout, and whenever it is stopped,
	it returns the best (smallest) change discovered so far.
	
	The search heuristic used by the top-level can have a crucial
	impact on performance. The approach implemented in \tool{} is
	general, in the sense that any search heuristic can be plugged
	in; and here, we consider and implement three such
	heuristics. The first is a random search, in which the top
	level randomly explores possible changes; this heuristic
	serves as a baseline. The second is a greedy search heuristic,
	in which the search always progresses in the direction that
	produces the most immediate gain. The third heuristic is a
	Monte Carlo Tree Search (MCTS)
	approach~\cite{BrPoWhLuCoRoTaPeSaCo12}, which attempts to
	balance between exploration of the search space and the
	explorations of regions already known to produce good
	solutions.
	
	The \tool{} tool will be made
        publicly available with the final version of this paper. It is
        designed in a modular fashion, so that additional search
        heuristics can be plugged in; it currently uses the Marabou
        DNN verification
        tool~\cite{KaHuIbJuLaLiShThWuZeDiKoBa19,WuOzZeIrJuGoFoKaPaBa20}
        as a backend, although other tools could be used as well. We
        used \tool{} to compare the different aforementioned heuristic
        strategies, and to compare our method to a single-layer
        modification method, with respect to the accuracy and minimal
        change size found. In our experiments, \tool{} achieved
        favorable results when compared to single-layer modification
        techniques. The greedy and MCTS heuristics both performed
        better than the random one; and while the greedy approach
        generally outperformed MCTS, there were cases where the latter
        proved superior.  Finally, we also used \tool{} to find
        three-layers modification to a network, as a proof-of-concept
        that demonstrates its ability to modify any number of layers
        simultaneously.

	The rest of this paper is organized as follows. In
	Section~\ref{sec:background} we provide the necessary
	background on DNNs and repairing DNNs with minimal
	modifications. In Section~\ref{sec:solution} we describe \tool{}'s
	algorithm for multi-layer modification in greater detail, and
	explain its different strategies for the heuristic
	search. Then, in Section~\ref{sec:implementation} we provide
	additional technical details on our implementation of
	\tool{}. We describe our experiments and results in
	Section~\ref{sec:experiments}. In
	Section~\ref{sec:relatedWork} we review relevant related work,
	and finally in Section~\ref{sec:conclusion} we conclude and
	describe our plans for future work.
	
	\section{Background}
	\label{sec:background}
	
	\subsubsection{Deep Neural Networks.}
	A deep neural network (a model) $N$ is comprised of $n$ layers,
	$L_1,\ldots,L_n$. Each layer $L_i$ is comprised of $s_i$ nodes, also
	called \emph{neurons}. The first layer, $L_1$, is the \emph{input
		layer}, and is used to provide the network with an input vector
	$v_1 \in \mathbb{R}^{s_1}$. The network is then evaluated by
	iteratively computing the assignment $v_i$ of layer $L_i$ for
	$i=2,\ldots,n$, each time using the assignment $v_{i-1}$ as part of
	the computation. Finally, the DNN computes the assignment $v_n$ of layer
	$L_n$, which is the \emph{output layer}. $v_n$ serves as the output of
	the entire neural network. Layers $L_2,\ldots,L_{n-1}$ are referred to
	as \emph{hidden layers}.
	
	Each assignment $v_i$ for $2\leq i \leq n$ is computed by multiplying
	$v_{i-1}$ by a real-valued \emph{weight vector} $\theta_i$, and
	applying a non-linear \emph{activation function} (except for the final
	output layer, where no activation function is applied). We use
	$\theta$ to denote the set of all weights
	$\theta= [\theta_2,\ldots,\theta_n]$, and use $N_\theta$ to refer to
	the function $N_\theta:\mathbb{R}^{s_1}\rightarrow\mathbb{R}^{s_n}$
	computed by $N$.  The weight vectors $\theta_i$ are key, and they are
	selected during the network's \emph{training phase}, which is beyond
	our scope here (see, e.g.,~\cite{GoBeCoBe16} for details).  Modern
	DNNs use various activation functions~\cite{NwIjWiGaMa18}; for
	simplicity, we restrict our attention here to the popular
	\emph{rectified linear unit} (\relu{}) function, defined as
	\[
	\relu{(x)}=\max\left(0,x\right),
	\]
	although our approach could be used with other functions as
	well.  When \relu{}s are used, the values $v_i$ of layer $L_i$
	are computed as $v_i=\relu{}(\theta_i\cdot v_{i-1})$, where
	the \relu{}s are applied element-wise.  We use the term
	\emph{network architecture} to refer to the number of layers
	in $N$, the size of each layer $s_i$, and the activation
	functions in use. Note that the network's weights are not
	considered part of the network's architecture.
	
	For a given point $x \in \mathbb{R}^{s_1}$, we refer to the assignment
	of the output layer $N_\theta(x)$ as the network's \emph{prediction}
	on $x$. A common class of DNNs are designed for the purpose of
	\emph{classification}, where the maximal entry of the prediction
	$N_\theta(x)$ indicates the \emph{label} to which $x$ is classified.
	In other words, the \emph{classification} of $x \in \mathbb{R}^{s_1}$
	as determined by $N_\theta$ is defined as $\arg \max
	N_\theta(x)$. Classification DNNs are useful, for example, for image
	recognition~\cite{SiZi14}, and are highly popular. When dealing with
	classification networks, we say that $N_\theta$ produces an erroneous
	output for $x$ if it classifies it differently than some given, ground-truth label $l$:
	\[
	\arg \max N_\theta(x) \neq l
	\]

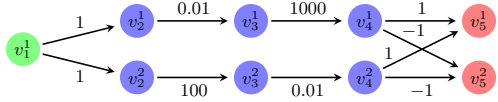
\begin{figure}[htp]
		\begin{center}
			\scalebox{0.75} {
				\def\layersep{2.0cm}
				\begin{tikzpicture}[shorten >=1pt,->,draw=black!50, node distance=\layersep,font=\footnotesize]
					
					\node[input neuron] (I-1) at (0,-2) {$v^1_1$};
					
					\node[hidden neuron] (H-1) at (\layersep,-1.5) {$v^1_2$};
					\node[hidden neuron] (H-2) at (\layersep,-2.5) {$v^2_2$};
					
					\node[hidden neuron] (H-3) at (2*\layersep,-1.5) {$v^1_3$};
					\node[hidden neuron] (H-4) at (2*\layersep,-2.5) {$v^2_3$};
					
					\node[hidden neuron] (H-5) at (3*\layersep,-1.5) {$v^1_4$};
					\node[hidden neuron] (H-6) at (3*\layersep,-2.5) {$v^2_4$};
					
					\node[output neuron] (O-1) at (4*\layersep, -1.5) {$v^1_5$};
					\node[output neuron] (O-2) at (4*\layersep, -2.5) {$v^2_5$};
					
					\draw[nnedge] (I-1) --node[above] {$1$} (H-1);
					\draw[nnedge] (I-1) --node[below] {$1$} (H-2);
					
					\draw[nnedge] (H-1) --node[above] {$0.01$} (H-3);
					\draw[nnedge] (H-2) --node[below] {$100$} (H-4);
					
					\draw[nnedge] (H-3) --node[above] {$1000$} (H-5);
					\draw[nnedge] (H-4) --node[below] {$0.01$} (H-6);
					
					\draw[nnedge] (H-5) --node[above] {$1$} (O-1);
					\draw[nnedge] (H-6) --node[above,pos=0.1] {$1$} (O-1);
					\draw[nnedge] (H-5)
					--node[above, pos=0.4] {$-1$} (O-2);
					\draw[nnedge] (H-6) --node[below] {$-1$} (O-2);
				\end{tikzpicture}
			}
			\captionsetup{size=small}
			\captionof{figure}{A toy DNN.}
			\label{fig:toyDnn}
		\end{center}
	\end{figure}

	A small, running example is depicted in Fig.~\ref{fig:toyDnn}. This
	toy DNN is comprised of five layers --- an input layer with a single
	node, three hidden layers with two nodes each, and an output layer
	with two nodes. The weight of each edge appears in the figure
	(a missing edge indicates a weight of $0$).  All activation functions
	in this example are \relu{}s. When the network is evaluated on input
	$v_1=[1]$, the assignment of the first hidden layer is $v_2=[1,1]$;
	the second hidden layer evaluates to $v_3=[0.01, 100]$;
	the third hidden layer evaluates to $v_4=[10,1]$; and
	finally, the output layer evaluates to $v_5=[11,-11]$.
	If we treat this DNN as a classification model,
	the classification of $x=1$ is $1$, as $11 = v_5^1 > v_5^2 = -11$.
	
	\subsubsection{Repairing DNNs with Minimal Modification.}
	
	For a given DNN $N_\theta:\mathbb{R}^{s_1}\rightarrow\mathbb{R}^{s_n}$
	with $n$ layers, and a finite set of points $S\subset\mathbb{R}^{s_1}$
	for which we know $N_\theta$ produces a wrong prediction, our goal is
	to change the network's weights $\theta$, so that its classification
	of $S$ becomes correct.
	
	We begin by formally defining the \emph{minimal modification problem}
	for classification networks (later, we extend this definition to other
	networks as well).  Let $N_\theta$ be a classification DNN, let $S$ be
	a set of inputs, and let $F$ be an oracle function 
	$F:\mathbb{R}^{s_1}\rightarrow\{1, \ldots, s_n\}$ which indicates the
	correct classification for each point $x \in S$. Our goal is to produce a modification to
	$\theta$, which we denote $\delta$, and obtain a new set of weights 
	$\theta' = \theta + \delta$, such that:
	\begin{equation}
		\label{eq_1}
		\forall x\in S. \quad \arg \max N_{\theta'}(x) = F(x)
	\end{equation}
	Observe that the architecture of $N$ is unchanged. Our goal is to find
	a $\delta$ that is \emph{minimal}, with the goal of preserving $N$'s
	behavior on points outside $S$. The magnitude of $\delta$ can be
	measured using any metric, such as the $L_1$ or $L_\infty$ norms.
	
	Using these definitions, the \emph{minimal modification problem} for
	classification DNNs is defined as follows:
	
	\begin{definition}
		\label{def:minModForClassification}
		\textbf{The Minimal Modification Problem for Classification Models.}
		Let $N_\theta:\mathbb{R}^{s_1}\rightarrow\mathbb{R}^{s_n}$ be a
		classification model with $n$ layers, and let
		$S\subset\mathbb{R}^{s_1}$ be a set of points. Let
		$F:S\rightarrow\{1, \ldots, s_n\}$ be an oracle
		function, which indicates the correct classification for each
		$x \in S$. Let $\norm{.}$ be some norm function.  The
		Minimal Modification Problem is:
		
		\[
		\begin{array}{ll@{}ll}
			\text{minimize} & \norm{\delta} &\\
			\text{subject to} & \arg \max N_{\theta'}(x) = F(x) & \quad \forall x \in S\\
			& \theta' = \theta + \delta
		\end{array}
		\]
		
	\end{definition}
	
\begin{figure}[t]
		\begin{center}
			\scalebox{0.75} {
				\def\layersep{2.0cm}
				\begin{tikzpicture}[shorten >=1pt,->,draw=black!50, node distance=\layersep,font=\footnotesize]
					
					\node[input neuron] (I-1) at (0,-2) {$v^1_1$};
					
					\node[hidden neuron] (H-1) at (\layersep,-1.5) {$v^1_2$};
					\node[hidden neuron] (H-2) at (\layersep,-2.5) {$v^2_2$};
					
					\node[hidden neuron] (H-3) at (2*\layersep,-1.5) {$v^1_3$};
					\node[hidden neuron] (H-4) at (2*\layersep,-2.5) {$v^2_3$};
					
					\node[hidden neuron] (H-5) at (3*\layersep,-1.5) {$v^1_4$};
					\node[hidden neuron] (H-6) at (3*\layersep,-2.5) {$v^2_4$};
					
					\node[output neuron] (O-1) at (4*\layersep, -1.5) {$v^1_5$};
					\node[output neuron] (O-2) at (4*\layersep, -2.5) {$v^2_5$};
					
					\draw[nnedge] (I-1) --node[above] {$1$} (H-1);
					\draw[nnedge] (I-1) --node[below] {$1$} (H-2);
					
					\draw[nnedge] (H-1) --node[above] {$0.01$} (H-3);
					\draw[nnedge] (H-2) --node[below] {$100$} (H-4);
					
					\draw[nnedge] (H-3) --node[above] {$1000$} (H-5);
					\draw[nnedge] (H-4) --node[below] {$0.01$} (H-6);
					
					\draw[nnedge] (H-5) --node[above] {$\overset{\hcancel[red]{1}}{\mathbf{-1.21}}$} (O-1);
					\draw[nnedge] (H-6)
					--node[above, pos=0.1] {$1$} (O-1);
					\draw[nnedge] (H-5)
					--node[above, pos=0.4] {$-1$} (O-2);
					\draw[nnedge] (H-6) --node[below] {$-1$} (O-2);
				\end{tikzpicture}
			}
			\captionsetup{size=small}
			\captionof{figure}{Minimal single-layer
				modification for the toy example of
				Fig.~\ref{fig:toyDnn}. The only changed
				layer is the output layer, where the weight
				of the edge $v_4^1 \rightarrow v_5^1$ was
				changed from $1$ to $-1.21$.  The size of
				the change (using the $L_1$ norm) is $2.21$.}
			\label{fig:toyDnnMinModSingleLayer}
		\end{center}
	\end{figure}
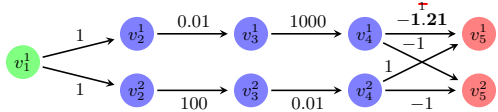

	We continue with our running example from
	Fig.~\ref{fig:toyDnn}. Recall that for input $x=1$, we get
	$v_5^1 = 11$ and $v_5^2 = -11$. Now assume that $S=\{1\}$, and
	that the desired classification for $x=1$ is actually
	$F(1)=2$. Thus, we need the network to satisfy that
	$v_5^1 < v_5^2$ when evaluated on $x=1$. We make an even
	stronger requirement, that $v_5^1 + \mu \leq v_5^2$, for some
	small $\mu>0$; this guarantees a small gap in the scores
	assigned to $v_5^1$ and $v_5^2$, and avoids draws. For this
	example, we set $\mu = 0.1$. Using the $L_1$ norm, the minimal
	single-layer modification that achieves the desired changes has size
	$2.21$, as depicted in
	Fig.~\ref{fig:toyDnnMinModSingleLayer}. With this change to
	the network, we get that $v_5^1=-11.1 < -11 = v_5^2$. However,
	if we allow changing two layers, we can actually achieve a
	\emph{smaller} minimal modification of size $2.11$, which is
	preferable because it has a smaller impact on the DNN's
	behavior. We will later return to this example in Section~\ref{sec:solution}.
	
	Definition~\ref{def:minModForClassification} is typically
	sufficient for classification DNNs, but it can be generalized
	to support arbitrary constraints on the DNN's outputs.  Let
	$N_\theta$ be a general DNN (not necessarily a classification
	DNN). For each point $x\in S$, we consider a matrix
	$A_x \in \mathbb{R}^{k_x\times s_n}$ and a vector
	$b_x \in \mathbb{R}^{k_x}$, where $k_x$ is the number of
	linear constraints on the output of the netowrk on $x$. The
	aim is to produce a modification to $\theta$, which we denote
	again with $\delta$, and get new weights
	$\theta' = \theta + \delta$, which satisfies:

	\begin{equation}
		\label{eq_1}
		A_x N_{\theta'}(x) \leq b_x
	\end{equation}

	Under this formulation we can express constraints such as
	``the first output of $N_{\theta'}$ on $x$ should satisfy
	$3 \leq N_{\theta'}(x) \leq 5$'', which could not be expressed
	using the previous formulation. This formulation subsumes the
	classification case.  Again notice that we keep the
	architecture of $N$ the same, and we only make modifications
	to $\theta$.  More formally, the minimal modification problem
	for the general case is defined as follows:
	
	\begin{definition}
		\label{def:minModGeneral}
		\textbf{The Minimal Modification Problem.} Let
		$N_\theta:\mathbb{R}^{s_1}\rightarrow\mathbb{R}^{s_n}$
		be a DNN model with $n$ layers, and let
		$S\subset\mathbb{R}^{s_1}$ be a set of points. For
		each point $x \in S$, let
		$A_x \in \mathbb{R}^{k_x\times s_n}$,
		$b_x \in \mathbb{R}^{k_x}$ be the output consraints of
		$N_\theta$ on $x$. Let $\norm{.}$ be some norm
		function.  The Minimal Modification Problem is:
		
		\[
		\label{eq_2}
		\begin{array}{ll@{}ll}
			\text{minimize} & \norm{\delta} &\\
			\text{subject to} & A_x N_{\theta'}(x) \leq b_x & \quad \forall x \in S\\
			& \theta' = \theta + \delta
		\end{array}
		\]
		
	\end{definition}		
	
	To the best of our knowledge, all previous approaches for
	solving the problems stated in
	Definitions~\ref{def:minModForClassification}
	and~\ref{def:minModGeneral} focused on finding a minimal
	modification for only a single layer of $N$. In contrast, in
	\tool{} we seek to solve the problem while allowing multiple
	layers of $N$ to be modified, as we discuss next.
	
	\section{\tool{}: Finding Multi-Layer DNN Changes}
	\label{sec:solution}
	
	The key idea incorporated into \tool{} is to reduce the multi-layer
	modification problem into a sequence of single-layer modification
	problems.  Specifically, given a DNN $N$ with $n$ layers
	$L_1,\ldots,L_n$ and a list of $k$ separation layer indices
	$1<i_1<\ldots<i_k<n$, we wish to partition the layers of $N$
	into $k+1$ sub-networks $N^0, N^1, \ldots, N^k$. Each sub-network is
	comprised of a subset of the original network's layers
	$L_1,\ldots,L_n$, as follows: sub-network $N^0$ is comprised of layers
	$L_1,\ldots,L_{i_1}$; sub-network $N^k$ is comprised of layers
	$L_{i_k},\ldots L_n$; and for each $1\leq j\leq k-1$, sub-network $N^j$
	is comprised of layers $L_{i_j},\ldots,L_{i_{j+1}}$.  Note that each
	pair of consecutive sub-networks $N^j$ and $N^{j+1}$ both contain
	layer $L_{i_{j+1}}$, which functions once as $N^j$'s output layer, and
	once as $N^{j+1}$'s input layer.  We refer to the shared layers
	$L_{i_1},\ldots,L_{i_k}$ as the \emph{separation layers}.
	
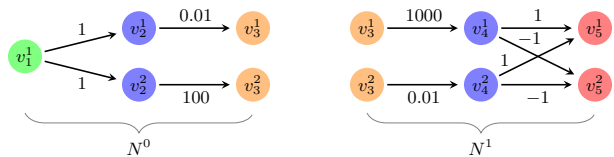
\begin{figure}[htp]
		\begin{center}
			\scalebox{0.75} {
				\def\layersep{2.0cm}
				\begin{tikzpicture}[shorten >=1pt,draw=black!50, node distance=\layersep,font=\footnotesize]
					
					\node[input neuron] (I1-1) at (0,-2) {$v^1_1$};
					
					\node[hidden neuron] (H1-1) at (\layersep,-1.5) {$v^1_2$};
					\node[hidden neuron] (H1-2) at (\layersep,-2.5) {$v^2_2$};
					
					\node[sep neuron] (O1-1) at (2*\layersep,-1.5) {$v^1_3$};
					\node[sep neuron] (O1-2) at (2*\layersep,-2.5) {$v^2_3$};
					
					\node[sep neuron] (I2-1) at (3*\layersep,-1.5) {$v^1_3$};
					\node[sep neuron] (I2-2) at (3*\layersep,-2.5) {$v^2_3$};
					
					\node[hidden neuron] (H2-1) at (4*\layersep,-1.5) {$v^1_4$};
					\node[hidden neuron] (H2-2) at (4*\layersep,-2.5) {$v^2_4$};
					
					\node[output neuron] (O2-1) at (5*\layersep, -1.5) {$v^1_5$};
					\node[output neuron] (O2-2) at (5*\layersep, -2.5) {$v^2_5$};
					
					\draw[nnedge] (I1-1) --node[above] {$1$} (H1-1);
					\draw[nnedge] (I1-1) --node[below] {$1$} (H1-2);
					
					\draw[nnedge] (H1-1) --node[above] {$0.01$} (O1-1);
					\draw[nnedge] (H1-2) --node[below] {$100$} (O1-2);
					
					\draw[nnedge] (I2-1) --node[above] {$1000$} (H2-1);
					\draw[nnedge] (I2-2) --node[below] {$0.01$} (H2-2);
					
					\draw[nnedge] (H2-1) --node[above] {$1$} (O2-1);
					\draw[nnedge] (H2-2) --node[above, pos=0.1] {$1$} (O2-1);
					\draw[nnedge] (H2-1) --node[above, pos=0.4] {$-1$} (O2-2);
					\draw[nnedge] (H2-2) --node[below] {$-1$} (O2-2);
					
					\draw[decorate,decoration={brace, amplitude=10, mirror}] (0, -3) -- (2*\layersep, -3) node[black,midway,yshift=-0.6cm] {$N^0$};
					\draw[decorate,decoration={brace, amplitude=10, mirror}] (3*\layersep, -3) -- (5*\layersep, -3) node[black,midway,yshift=-0.6cm] {$N^1$};
				\end{tikzpicture}
			}
			\captionsetup{size=small}
			\captionof{figure}{Splitting a network along a
				separation layer.}
			\label{fig:toyDnnSplitting}
		\end{center}
	\end{figure}

	We apply this partitioning to our running example, as depicted in Fig.~\ref{fig:toyDnnSplitting}. There, we split the DNN into
	two sub-networks $N^0$ and $N^1$, with the original $L_3$ layer
	serving as the only separation layer.  Observe that the input layer of $N^0$ is the input layer of the original network, and that the output layer of $N^1$ is the output layer of the original network. Indeed, if we were to evaluate $N^0$ on some input $x$, and then feed its output as the input to $N^1$, then $N^1$'s output would match the output of the
	original network when evaluated on $x$.

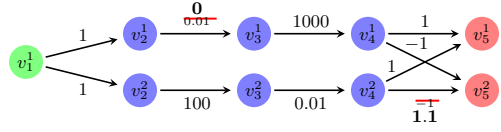
\begin{figure}[htp]
		\begin{center}
			\scalebox{0.75} {
				\def\layersep{2.0cm}
				\begin{tikzpicture}[shorten >=1pt,draw=black!50, node distance=\layersep,font=\footnotesize]
					
					\node[input neuron] (I1-1) at (0,-2) {$v^1_1$};
					
					\node[hidden neuron] (H1-1) at (\layersep,-1.5) {$v^1_2$};
					\node[hidden neuron] (H1-2) at (\layersep,-2.5) {$v^2_2$};
					
					\node[hidden neuron] (H2-1) at (2*\layersep,-1.5) {$v^1_3$};
					\node[hidden neuron] (H2-2) at (2*\layersep,-2.5) {$v^2_3$};
					
					\node[hidden neuron] (H3-1) at (3*\layersep,-1.5) {$v^1_4$};
					\node[hidden neuron] (H3-2) at (3*\layersep,-2.5) {$v^2_4$};
					
					\node[output neuron] (O1-1) at (4*\layersep, -1.5) {$v^1_5$};
					\node[output neuron] (O1-2) at (4*\layersep, -2.5) {$v^2_5$};
					
					\draw[nnedge] (I1-1) --node[above] {$1$} (H1-1);
					\draw[nnedge] (I1-1) --node[below] {$1$} (H1-2);
					
					\draw[nnedge] (H1-1) --node[above] {$\underset{\hcancel[red]{0.01}}{\mathbf{0}}$} (H2-1);
					\draw[nnedge] (H1-2) --node[below] {$100$} (H2-2);
					
					\draw[nnedge] (H2-1) --node[above] {$1000$} (H3-1);
					\draw[nnedge] (H2-2) --node[below] {$0.01$} (H3-2);
					
					\draw[nnedge] (H3-1) --node[above] {$1$} (O1-1);
					\draw[nnedge] (H3-2)
                                        --node[above, pos=0.1] {$1$} (O1-1);
					\draw[nnedge] (H3-1)
                                        --node[above, pos=0.4] {$-1$} (O1-2);
					\draw[nnedge] (H3-2) --node[below] {$\overset{\hcancel[red]{-1}}{\mathbf{1.1}}$} (O1-2);
				\end{tikzpicture}
			}
			\captionsetup{size=small}
			\captionof{figure}{The two-layer modification found using \tool{}.}
			\label{fig:toyDnnMinModMultiLayer}
		\end{center}
	\end{figure}

	Next, we wish to modify $N^0$ and $N^1$, and then combine these
	modifications into a modification of the original network.  Let
	$S=\{1\}$, i.e.  $x=1$ is our only misclassified input, and let us
	require that $x$ be classified as class $2$. In other words, we wish
	$N(1)$ to produce output values for which $v_5^1 + \mu \leq v_5^2$ for
	some small $\mu>0$.  \tool{} begins by deciding on a change of values
	for the neurons of the separation layer, $v^1_3$ and $v^2_3$.  In the
	original evaluation of the network on $x=1$, we got $v_3^1=0.01$ and
	$v_3^2=100$. Let us require that $v_3^1$'s value be changed to $0$,
	and that $v_3^2$'s value remains unchanged. This requirement
	translates into two single-layer modification queries: for $N^0$,
	\tool{} will require that on input $x=1$, the outputs be $[0,100]$;
	and for $N^1$, \tool{} will require that on input $[0,100]$, the
	network's outputs satisfy $v_5^1 + \mu \leq v_5^2$. Both these
	single-layer modification queries can be solved using a black-box
	modification procedure; for example, here, if we assume again that
	$\mu = 0.1$, a possible modification is to change the weight of edge
	$v_2^1\rightarrow v_3^1$ to $0$ in $N^0$, and to change the weight of
	edge $v_4^2 \rightarrow v_5^2$ to $1.1$ in $N^1$. Applying both of
	these changes to the original network produces a modification of size
	$2.11$ (using the $L_1$-norm), which results in the desired behavior
	for $x=1$; indeed, after applying this change, we get that
	$1 = v_5^1 < v_5^2 = 1.1$. The modified network is depicted in
	Fig.~\ref{fig:toyDnnMinModMultiLayer}.  Observe that this change is
	minimal for our particular selection of a separation layer index and
	the ensuing selection of changes to the separation layer's assignment;
	but it is not necessarily globally minimal, as a different choice of
	separation index or assignment could result in smaller changes.

	The example described above is generalized into \tool{}'s full
	algorithm, which appears as Algorithm~\ref{alg:solution}.  For
	simplicity of presentation, Algorithm~\ref{alg:solution} handles the
	classification model case from
	Definition~\ref{def:minModForClassification}; \tool{} actually
	supports the more general case from
	Definition~\ref{def:minModGeneral}, and the implemented algorithm is
	very similar to the one given here.  Algorithm~\ref{alg:solution}
	takes as input the DNN $N$ in question, the set of misclassified
	points $S$ and the oracle function $F$ that describes these points
	desired classification; the separation indices $I=\{i_1,\ldots,i_k\}$
	indicating how the network is to be broken down into sub-networks, in
	which only a single layer will be changed; and a timeout value
	$T$. The algorithm then begins its heuristic search for a minimal
	change to the network that brings about the desired changes.

	\begin{algorithm}
		\caption{The \tool{} Algorithm (For Classification Networks)}
		\hspace*{\algorithmicindent} \textbf{Input:} DNN $N$, set of input points $S=\{x_1,\ldots,x_n\}$, oracle function $F$, separation indices $I=\{i_1,\ldots,i_k\}$, timeout $T$ \\
		\hspace*{\algorithmicindent} \textbf{Output:} A repaired DNN $N'$ with the same architecture as $N$
		\begin{algorithmic}[1]
			\For{$j=1 \ldots n$} \label{line:startComputeSepLayers}
			\State $v^j_{i_1},\ldots,v^j_{i_k} \gets N(x_j)$  \Comment{Compute
				the separation layers' assignments} 
			\EndFor \label{line:endComputeSepLayers}
			\State $N^0,\ldots,N^k \gets \Call{Split}{N,I}$ \label{line:split}
			\State $\bc{}\leftarrow \bot, \bestCost{}\leftarrow \infty$ \label{line:initCost}
			
			\While{timeout $T$ not exceeded} \label{line:startWhile}
			
			\For{$l=1\ldots k$}
			\State $c_l \gets \Call{ProposeChange}$ \label{line:proposeChange}
			\For{$j=1\ldots n$} \label{line:startModifySepLayers}
			\State $v'^j_{i_l} \gets v^j_{i_l} + c_l$ \Comment{Select new assignments for the separation layers}
			\EndFor \label{line:endModifySepLayers}
			\EndFor  
			
			\State $N'^0, cost_0 \gets \Call{SingleLayerModification}{N^0,
				\langle x_1,v'^1_{i_1} \rangle,\ldots, \langle x_n, v'^n_{i_1}\rangle}$ \label{line:modifyN0}
			\For{$l=1\ldots k-1$}
			\State $N^l, cost_l \gets \Call{SingleLayerModification}
			{N'^l, \langle v'^1_{i_l}, v'^1_{i_{l+1}}\rangle,\ldots, \langle v'^n_{i_l}, v'^n_{i_{l+1}}\rangle}$ \label{line:modifyNl}
			\EndFor
			\State $N'^k, cost_k \gets \Call{SingleLayerModification}{N^k, \langle v'^1_{i_k},F(x_1)\rangle,\ldots,\langle v'^n_{i_k},F(x_n)\rangle }$ \label{line:modifyNk}
			\State  $\cost{}\gets \Call{TotalCost}{\cost{}_0,\ldots,\cost{}_k}$ \label{line:totalCost}
			\If {$\cost{}< \bestCost{}$} \label{line:startSaveBest}
			\State $\bestCost{}\gets \cost{}$
			\State $\bc\leftarrow\langle N'^0,\ldots,N'^k\rangle$
			\EndIf \label{line:endSaveBest}
			\EndWhile \label{line:endWhile}
			\State \Return $\langle \bestCost, \Call{Combine}{\bc} \rangle $
		\end{algorithmic}
		\label{alg:solution}
	\end{algorithm}
	
	First, in
	Lines~\ref{line:startComputeSepLayers}--\ref{line:endComputeSepLayers},
	the algorithm evaluates the assignments of the separation layers, for
	each input point in $S$. Then, in Line~\ref{line:split}, the algorithm
	constructs the sub-networks $N^0,\ldots,N^k$, according to the
	separation indices.  Recall that our algorithm is anytime, i.e., always
	maintains the best modification discovered so far; this modification,
	and its cost (i.e., its distance from the original network according
	to the distance metric in use) is stored in the variables initialized
	in Line~\ref{line:initCost}. The algorithm then begins running in a
	loop until exhausting its timeout value.
	
	In every iteration of its main loop, the algorithm begins
	(Lines~\ref{line:proposeChange}--\ref{line:endModifySepLayers}) by
	selecting a modified assignment for each separation layer $L_{i_l}$
	for $1\leq l\leq k$. This modification is selected by the place-holder
	function \textsc{ProposeChange()}; this function is where the heuristic search used in the search level of \tool{} comes into play. We
	discuss these heuristics in detail in
	Section~\ref{subsection:searchLevel}. Then, in
	Lines~\ref{line:modifyN0}--\ref{line:modifyNk}, the algorithm computes
	for each of the sub-networks $N^0,\ldots,N^k$ the minimal,
	single-layer changes required to bring about the global changes
	selected by the search level.  These changes are computed by repeated
	invocations of the \textsc{SingleLayerModification()} function, which
	is again a place-holder function that represents the single-layer
	modification level of \tool{}; we describe it in more depth in
	Section~\ref{subsection:singleLayerMod}.  This function takes as input
	a DNN, and a list of pairs of input points and their desired outputs;
	and returns the modified DNN, and the modification's cost.\footnote{It
		may be possible that an invocation of
		\textsc{SingleLayerModification()} fails because no change is
		possible that obtains the desired results. Whenever this happens,
		\tool{} continues to the next iteration, exploring a different
		change to the separation layers' values. This situation is
		theoretically possible, but did not occur in our experiments.} In
	Line~\ref{line:modifyN0}, we use \textsc{SingleLayerModification()} to
	modify $N^0$: we required that the original input points
	$x_1,\ldots,x_n$ produce outputs that match the selected modified
	assignments $v'^1_{i_1},\ldots,v'^n_{i_1}$ of $L_1$. In
	Line~\ref{line:modifyNl}, \textsc{SingleLayerModification()} is used
	to modify each of the $N^1,\ldots,N^{k-1}$ sub-networks, so that each
	sub-network produces as output the input selected for its successor.
	Finally, in Line~\ref{line:modifyNk}, the last sub-network $N^k$ is
	modified, so that it produces outputs that match the oracle's
	predictions on the original input points.
	
	The single-layer modification procedures invoked for $N^0,\ldots,N^k$
	each return the modified sub-networks $N'^0,\ldots,N'^k$, and the
	respective costs of the modifications $cost_0,\ldots,cost_k$. The
	total modification cost for the complete network is then computed by
	the \textsc{TotalCost()} function in Line~\ref{line:totalCost}, whose
	implementation depends on the norm used for measuring distance;
	for example, in the case of $L_1$ norm, it returns the sum of its inputs;
	for $L_\infty$, it returns the maximal input; etc. The modified
	sub-networks with the lowest total cost found so far, along with 
	the cost itself, are saved in
	Lines~\ref{line:startSaveBest}--\ref{line:endSaveBest}.
	
	The algorithm halts when the provided timeout is exhausted, and it
	then returns the complete modified network with the best modifications
	found so far, and the cost of that modification. The re-assembling of
	the complete modified network is performed by the function
	\textsc{Combine()}, whose implementation is omitted for brevity.

	\subsubsection{Soundness and Completeness.}	
	Assuming that the \textsc{SingleLayerModification()} is sound --- for
	example, if it is implemented using a sound DNN
	verifier~\cite{GoAdKeKa20} --- any modification returned by our tool
	will indeed correct the global DNN behavior on the input set $S$. In
	that sense, \tool{} is sound. It is, however, generally incomplete;
	there are infinitely many modifications that can be attempted for the
	separation layers, and it is infeasible to try them all. This is our
	motivation for introducing the timeout mechanism and making the
	algorithm anytime; and indeed, the algorithm is not guaranteed to
	return the smallest change possible. It does, however, attempt to
	minimize the change based on search heuristics that we discuss next.

	\subsection{The Search Level}
	\label{subsection:searchLevel}
\begin{wrapfigure}[19]{r}{5cm}
		\vspace{-1.0cm}
		\begin{center}
			\includegraphics[width=50mm]{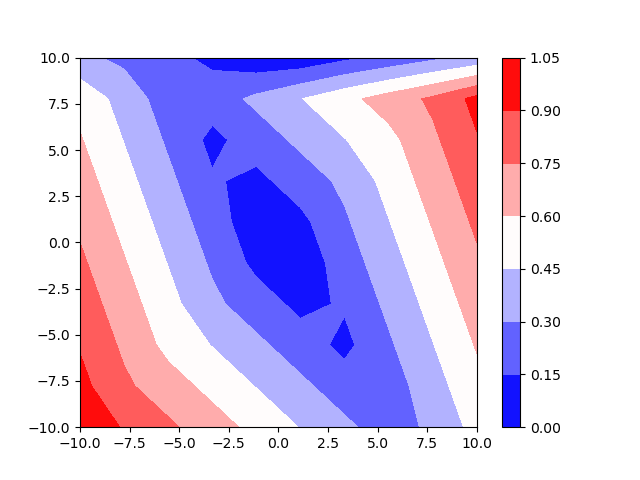}		
		\end{center}
		\caption{The cost function for a small DNN, with a single
			separation layer with 2 neurons. The X and Y axes represents
			the change for each neuron, and the color represents the
			size of the minimal modification achieved. The function is
			not convex.}
		\label{fig:searchSpace}
	\end{wrapfigure}

	Algorithm~\ref{alg:solution} considers an infinite space of possible
	changes to the values of the separation layers, each time selecting a
	single possible change and computing its cost
	(Line~\ref{line:proposeChange} of the Algorithm). For a single
	separation layer with $n$ neurons, the search space is
	$\mathbb{R}^{n}$ in its entirety, and the problem is compounded when
	multiple separation layers are involved. To exacerbate matters even
	further, the computed cost function for possible changes need not be
	convex; see Fig.~\ref{fig:searchSpace} for an illustration.
	
	To circumvent this difficulty, we first define the following grid, parameterized by a step size $\epsilon$:
	\[
	\mathcal{G}_\epsilon= \{ v = \langle \alpha_1\cdot \epsilon, \alpha_2\cdot
	\epsilon, \ldots, \alpha_n\cdot \epsilon\rangle\ |\ \alpha_i \in \mathbb{Z}\}
	\]
	Each point in the grid represents a single, possible change for a
	separation layer, and the
	discretization allows us to better handle the search space. Naturally,
	this comes at the cost of possibly overlooking better changes that do
	not coincide with the grid, but this can be mitigated by making the
	grid dense (picking a smaller $\epsilon$).  The grid's origin,
	i.e., point $0^n\in \mathbb{R}^n$, corresponds to no change at all to
	the separation layer; and points that are very far away from the
	origin are likely to represent significant changes to the DNN.
	
	Despite the discretization, the grid is still infinite and
	multi-dimensional, and so \tool{} implements three search
	heuristics: \emph{random search}, \emph{greedy search} and
	\emph{Monte-Carlo Tree Search (MCTS)}. Each of these
	heuristics can be regarded as a possible implementation of the
	\textsc{ProposeChange()} method from Algorithm~\ref{alg:solution}.  We
	next elaborate on each of them.
	
	\subsubsection{Random Search.} This heuristic performs a uniform
	random search over $\mathcal{G}_\epsilon$. Specifically, it samples a
	grid point uniformly at random, and that point constitutes the proposed
	change to the separation layer. We treat this simple heuristic as a
	baseline, to which the more sophisticated heuristics are compared.
	
	\subsubsection{Greedy Search.} The motivation for this heuristic is
	that the optimal grid point is likely not far away from the origin (as
	far away points likely correspond to significant changes to the
	network). Thus, we start from the grid's origin as our current change,
	and at each iteration, consider the grid points that are immediate
	neighbors of our current points --- that is, points obtained by adding
	or subtracting $\epsilon$ from one of the coordinates of the current
	point. We then compute the costs associated with each of these points,
	and pick the cheapest one as our new current point.
	
	More formally, if $g_0\in \mathcal{G}_\epsilon$ is our current search point, we observe all points $g\in\mathcal{G}_\epsilon$ such that $\lVert g_0-g\rVert_{L_1}=\epsilon$, invoke the
	\textsc{SingleLayerModification()} with appropriate paremeters to compute the cost of each $g$, and update $g_0$ to be the $g$ that obtained the lowest cost.
	
	\subsubsection{Monte Carlo Tree Search.}
	
	The aforementioned greedy approach can be regarded as an attempt to
	optimize \emph{exploitation}: whenever a good ``direction''
	on the grid is discovered, we follow that direction. A natural concern
	is that such an approach might lead to local minima, and fail to
	detect cheaper changes that can only be reached via grid points with
	higher costs (recall that the cost function is not necessarily
	convex). To balance the greedy approach's exploitation with
	\emph{exploration} for detecting possibly better changes, we employ a
	\emph{Monte Carlo Tree Search} (\emph{MCTS})
	heuristic~\cite{BrPoWhLuCoRoTaPeSaCo12}.  We give here a short
	overview of this approach; see~\cite{BrPoWhLuCoRoTaPeSaCo12} for a
	more in-depth review.
	
	MCTS is a heuristic search algorithm over a discrete set of actions, with the goal of selecting the most promising move based on simulations. It has recently been shown quite successful in multiple application domains, most notably in board games such as Go~\cite{Fu16}. The search is conducted on a \emph{search tree}, where
	each node represents a state. The root node of the search tree
	represents the initial state, and a child of a node represents another state that can be reached by performing a single action. Initially,  the entire search tree is yet \emph{unexplored}; and the algorithm
	iteratively explores additional parts thereof, one node in every
	iteration. In our setting, each node of the search tree is a grid point; and the possible moves include moving to one of the adjacent grid points (similarly to the greedy approach).
	
	In each iteration, MCTS performs \emph{simulations} in order to decide which unexplored node to visit next. Specifically, these simulations allow MCTS to compute a cost for each of the candidate nodes, and
	then pick the candidate with the lowest cost as the next node to visit.
	
	More concretely, each MCTS iteration consists of 4 steps:
	\begin{enumerate}
		\item \emph{Selection}: one of the nodes at each level in the explored portion of
		tree is selected, according to some policy, until reaching a leaf node. A common policy, also
		used in \tool{}, is the \emph{upper
			confidence bound} (\emph{UCB}) policy. The policy's details are
		beyond our scope here; see~\cite{BrPoWhLuCoRoTaPeSaCo12} for
		additional details.
		
		\item \emph{Expansion}: one of the unexplored children of the leaf node from Step 1 is selected randomly.
		
		\item \emph{Simulation}: one or more simulations are carried out for
		the node selected in Step 2. Each simulation explores deeper into
		the search sub-tree rooted at the new node until reaching a predefined
		tree depth, by picking a child randomly in each level of
		the sub-tree. When the simulation arrives at the last node, it
		computes a cost value that takes into account all the steps that led
		from the node picked at Step 2 to the final node that was reached.
		
		\item \emph{Backpropagation}: the cost computed in each simulation is
		back-propagated through all the nodes in the path leading back up to
		the root. Each node aggregates the costs of simulations of paths
		containing it, and the aggregated cost is used for Step 1 in the
		next iteration of MCTS.
		
	\end{enumerate}
	
	After reaching a predefined number of iterations, the unexplored node
	that has obtained the lowest cost so far is chosen as the next move.
	
	In our implementation of the MCTS search heuristic, every invocation
	of \textsc{ProposeChange()} for a given separation layer
	$L_j$ runs the MCTS algorithm, which in turn performs a predefined
	number of sub-iterations. The root of the search tree represents the
	current change to the assignment of $L_j$, and a move to a child node
	represents a single step along the grid. Consequently, for each tree
	node of the search tree in the MCTS algorithm, there are $2s_j + 1$
	child nodes (including the option to not take a step at all). The
	simulation step of MCTS includes, in our case, dispatching
	single-layer modification queries.
	
	\subsection{The Single-Layer Modification level}
	\label{subsection:singleLayerMod}
	As part of its operation, our algorithm needs to dispatch numerous
	queries of single-layer modifications in DNN (the
	\textsc{SingleLayerModification()} calls in
	Algorithm~\ref{alg:solution}). In each of these queries, the
	sub-network in question has specific inputs, for which certain output
	constraints need to hold --- either the outputs need to classify the
	inputs as a certain label (for the last sub-network), or they need to
	take on exact, predetermined values (for all other
	sub-networks). Solving such queries has been studied before, and as
	part of our solution, we propose to use existing techniques and tools
	as a backend. In our implementation (described in greater detail
	later), we used the approach proposed by Goldberger
	et. al~\cite{GoAdKeKa20}.
	
	\section{Implementation}
	\label{sec:implementation}
	We implemented Algorithm~\ref{alg:solution} and the aforementioned
	search heuristics in the new \tool{} tool. \tool{} is implemented
	as a Python $3.7.3$ module, and uses TensorFlow-Keras $2.3$ as a backend
	for representing DNNs. We attempted to design \tool{} in a modular
	fashion, in order to easily allow the future addition of new search
	heuristics in the search level, as well as additional backend engines
	for dispatching single-layer modification queries.
	
	The main class of \tool{} is the abstract
	\emph{NetworkCorrectionMethod} class. It defines the interfaces and
	methods that a subclass must implement in order to fit the mold
	defined by Algorithm~\ref{alg:solution}. Specifically, the class
	defines the following methods:
	
	\subsubsection{\_\_init\_\_(DNN $N$, $[x_1,\ldots,x_n]$, $[o_1,\ldots,o_n]$):} a constructor for the inheriting
	class. It takes as input a TensorFlow-Keras DNN, a list of input
	points as NumPy arrays, and a list of output constraints for each
	point. Each output constraint is a list of 2 items: a NumPy array $A$
	and a NumPy vector $b$, and the output $y$ of the corresponding point
	should satisfy $Ay\leq b$ (per Definition~\ref{def:minModGeneral}).
	
	\subsubsection{correct\_network()} the main entry point for the
	inheriting class, responsible for running the correction procedure for
	the DNN and constraints provided through the constructor. Its
	implementation depends on the heuristic search method and the single-layer modification method chosen. Returns \emph{True} if a
	modification to the network was found, or \emph{False} otherwise.
	
	\subsubsection{get\_corrected\_network():} this method is invoked after
	\emph{correct\_network()}, and returns the corrected network as a tensorflow-keras model.
	
	\subsubsection{get\_minimal\_change():} a method called after
	\emph{correct\_network()}, which returns the list of the changes found during the modification process, for each changed layer.
	
	\subsubsection{get\_changed\_layers():} a method called after
	\emph{correct\_network()}, which returns a list of layer indices of the layers changed during the modification process.
	
	\medskip
	
	Our implementation of \tool{} includes multiple instantiations of the
	\emph{NetworkCorrectionMethod} class that implement the heuristics
	defined in Section~\ref{sec:solution}. Specifically, class
	\emph{NetworkCorrectionTwoLayersUniform} implements the random search
	heuristic; the core of the implementation appears in the
	\emph{correct\_network()} method. Similarly, class
	\emph{NetworkCorrectionTwoLayersGreedy} implements the greedy search
	approach; and its core is again in method
	\emph{correct\_network()}. Finally, the MCTS approach is implemented
	in classes \emph{NetworkCorrectionTwoLayersTreeSearch} and
	\emph{MCTS}. Class \emph{MCTS} controls the various configurable
	parameters of the search, such as the step size, the number of
	simulations per iteration, and the maximal depth of the search
	tree. All three grid search heuristics are currently linked to the
	Marabou DNN verification as the single-layer change backend; this
	connection is implemented in class \emph{MarabouRunner}.
	
	\begin{table}[htp]	
		\centering	
		\begin{tabularx}{\textwidth} {	
				| >{\centering\arraybackslash}p{0.8cm}|X
				| >{\centering\arraybackslash}X	
				| >{\centering\arraybackslash}X	
				| >{\centering\arraybackslash}X	
				| >{\centering\arraybackslash}X	
				| >{\centering\arraybackslash}X	
				| >{\centering\arraybackslash}X	
				| >{\centering\arraybackslash}X	
				| >{\centering\arraybackslash}X |}	
			\hline	
			Exp. \# & Search Strategy & Number of input points & Average Change & Minimal Change & Maximal Change & Average Accuracy & Minimal Accuracy & Maximal Accuracy \\	
			
			\hline\hline	
			
			\multirow{6}{0.5em}{1} & Random & \multirow{3}{0.5em}{1} & 0.1520 & 0.0615 & 0.4922 & 0.6865 & 0.1916 & 0.9308 \\	
			& Greedy & & 0.0133 & 0.001 & 0.0566 & 0.943 & 0.7971 & 0.9576 \\	
			& MCTS & & 0.0139 & 0.001 & 0.0566 & 0.943 & 0.7971 & 0.9576 \\
			
			\cline{2-9}		
			
			& Random & \multirow{3}{0.5em}{2} & 0.197 & 0.0791 & 0.4775 & 0.6302 & 0.2563 & 0.9161 \\	
			& Greedy & & 0.0463 & 0.0058 & 0.1435 & 0.9245 & 0.7417 & 0.9598 \\	
			& MCTS & & 0.0478 & 0.0058 & 0.1484 & 0.9261 & 0.7398 & 0.9594 \\	
			
			
			
			\hline\hline	
			
			\multirow{4}{0.5em}{2} & Greedy & \multirow{2}{0.5em}{1} & 0.0305 & 0.0029 & 0.1699 & 0.9397 & 0.9565 & 0.5856 \\	
			& 1-Layer & & 0.0307 & 0.0029 & 0.1875 & 0.9394 & 0.585 & 0.9562 \\
			
			\cline{2-9}	
			
			& Greedy & \multirow{2}{0.5em}{2} & 0.0459 & 0.0039 & 0.2041 & 0.9178 & 0.3124 & 0.9576 \\	
			& 1-Layer & & 0.0464 & 0.0039 & 0.208 & 0.9163 & 0.3124 & 0.9576 \\
			
			\hline\hline	
			
			3 & Greedy-3 & 1 & 0.25097 & 0.25097 & 0.25097 & 0.886 & 0.886 & 0.886 \\	
			
			\hline	
			
		\end{tabularx}	
		\caption{Results of experiments. The 1-Layer search strategy stands for a single-layer modification process. Greedy-3 stands for three-layer-modification using the greedy heuristic search.}	
		\label{table:results}	
	\end{table}	
	
	\section{Evaluation}
	\label{sec:experiments}
	
	\subsubsection{Setup.}
	
	We used \tool{} to evaluate the usefulness of our modification	
	approach. Specifically, we experimented with a DNN trained on the MNIST	
	dataset for digit recognition~\cite{Le98}. The dataset contains 70,000	
	handwritten digit images with $28\times28$ pixels, split into	
	a training set of 60,000 images, and a test set of 10,000 images. We	
	trained a network $N$ comprised of 8 layers: an input layer of size 784	
	neurons, six hidden layers, each of size 20 neurons, and an output layer	
	with ten neurons. The hidden layers all used the \relu{} activation	
	function. 
	We then used network $N$ to conduct three kinds of experiment
	(all conducted with the $L_\infty$-norm):
	\begin{inparaenum}[(i)]	
		\item comparing search heuristics: an experiment where we used \tool{} to	
		find two-layer modifications for $N$, using each of the three	
		heuristic search strategies discussed in Section~\ref{sec:solution};	
		\item comparing multi-layer and single-layer modifications: here, we	
		used \tool{} to search for repairs for $N$ that modified either a	
		single layer or two layers, in order to evaluate the necessity of	
		modifying multiple layers; and	
		\item three-layer repairs: we attempted to repair $N$ by modifying	
		three layers, to demonstrate \tool{} ability to repair the network	
		by changing any number of layers.	
	\end{inparaenum}	
	Below we provide additional information on each of the experiments,
	and their
	results are summarized in Table~\ref{table:results}.
	
	\begin{wrapfigure}[13]{r}{5cm}
		\vspace{-1.4cm}
		\begin{center}
			\includegraphics[width=50mm]{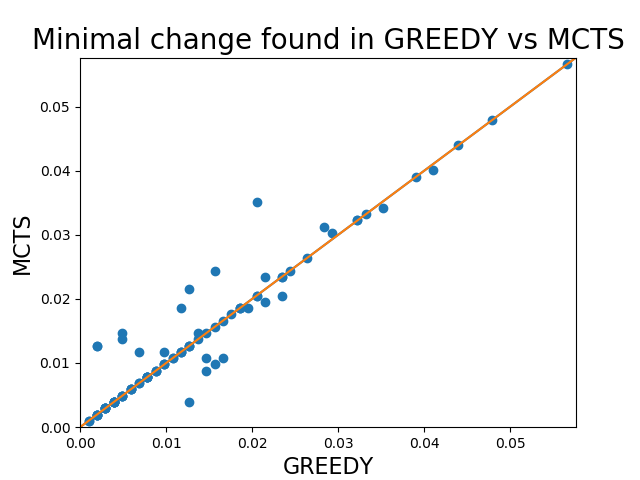}		
		\end{center}
		\caption{Minimal modification size achieved by the Greedy and MCTS heuristic strategies in Experiment 1.}
		\label{fig:figGreedyVsMctsChange}
	\end{wrapfigure}
	
	\subsubsection{Experiment 1: comparing search heuristics.}	
	We used \tool{} in each of the three search method configurations, to solve:
	\begin{inparaenum}[(i)]
		\item 100 benchmarks where $N$ was modified to repair its output on 1 input
		point; and
		\item another 100 benchmarks with repair on 2 input points.
	\end{inparaenum}
	In all experiments, we split $N$ into two sub-networks along its
	fourth hidden layer, with $\epsilon=0.5$ as the grid parameter; and
	the timeout value was set to 1000 seconds.  In experiments
	involving two input points, we expedited the process by restricting changes solely to the final layer of each sub-network.  The results are
	summarized in Table~\ref{table:results}. Both the Greedy and MCTS
	strategies significantly outperform the uniform random search
	heuristic, achieving higher accuracy and smaller change size. The Greedy
	and MCTS heuristics are relatively equal in their performance, with
	each strategy outperforming the other in some cases.

	\subsubsection{Experiment 2: comparing multi- and single-layer	
		modifications.}	
	Here, we configured \tool{} to use the greedy search heuristic,
	and used it to solve: \begin{inparaenum}[(i)]
		\item 2000 minimal modification queries where where a
		single input point had to be corrected; and
		\item another 2000 minimal modification queries with repair on 2 input points.
	\end{inparaenum} We ran each query
	once, looking for a one-layer minimal modification, and once
	searching for a two-layer modification. As before, we set
	$\epsilon=0.5$ and a timeout value of 1000 seconds (for both
	methods).  To expedite the experiments, we allowed the
	single-layer method to modify only the final layer of the
	network~\cite{GoAdKeKa20}, and the two-layers greedy method to
	modify the last layer of each of the two sub-networks.
	Table.~\ref{table:results} shows the superior performance of
	the two-layers greedy method over the single-layer method;
	although the single-layer modification method was usually able
	to find its minimal modification within a minute, while 
	the two-layers greedy method took longer. This is not surprising, as
	the single-layer modification problem is significantly easier
	computationally~\cite{GoAdKeKa20}.
	
	\subsubsection{Experiment 3: three-layer repairs.} In the final experiment, serving as a proof-of-concept, we used \tool{} to find a
	three-layer modification for $N$. We ran this experiment once,
	with \tool{} configured to use the greedy search heuristic on a
	single input point. We used a step size of $\epsilon=0.5$. The
	timeout value was set to 3600 seconds, and
	Table~\ref{table:results} depicts the results. The search space
	when changing three layers is significantly more complex than in the
	previous experiments, and so it is not surprising that \tool{}
	was only able to discover changes that were larger than
	before. As we continue to improve our search heuristics, and as
	the underlying verification engines continue to improve, the
	scalability of \tool{} will also improve.
	
	\section{Related Work}
	\label{sec:relatedWork}
	
	The need to modify existing DNNs in order to correct them naturally
	arises as part of the DNN life cycle, and has been a topic of interest
	in the wider machine learning community. Most existing approaches are
	heuristic in nature: for example, one approach is to iteratively apply
	Max-SMT solvers in search for changes to the DNN~\cite{SoTh19};
	another is to use reachability analysis to enrich the training
	data~\cite{YaYaTrHoJoPo21}; and yet another approach is to
	heuristically identify ``problematic'' neurons and modify
	them~\cite{DoSuWaWaDa20}. A common
	property of most of these approaches is that, in contrast to
	verification-based approaches, they provide no formal guarantees about
	the minimality of the fixes that they produce.
	
	Another approach for modifying the behavior of an existing DNN is to
	augment it with additional, non-DNN components that can override its
	output in certain cases. This has been attempted using, e.g., decision
	trees~\cite{KaLe18,KaFu18} and scenario objects~\cite{Ka20,KaEl21}. A
	different technique is to transform the DNN into another object, which
	is simpler to repair: for example, a pair of DNNs, in which one
	determines the weights and another the activation
	functions~\cite{SoTh21}; or a DNN with a self-repairing output
	layer~\cite{LeFrMaFrPaPa21}.  Our technique is separated from these
	approaches by the fact that the repaired artifact that it produces is
	a standard DNN, and is thus directly compatible with existing tools
	and infrastructure.
	
	The approach that we take here, namely the application of DNN
	verification technology in order to find minimal
	modifications, has already received some attention.
	The approach that most closely resembles our own is the one
	proposed by Goldberger et al.~\cite{GoAdKeKa20}; and a related
	approach has also been proposed by Usman et
	al.~\cite{UsGoSuNoPa21}.  However, these approaches are
	limited to modifying a single layer of the DNN in question,
	whereas the novelty of our approach is in enabling the
	simultaneous modification of multiple layers.
	
	The technique proposed here uses a DNN verification engine as
        a black-box.  DNN verification is an active research field,
        with many available tools and techniques. These include
        SMT-based
        approaches~\cite{HuKwWaWu17,KaBaDiJuKo17,KaHuIbJuLaLiShThWuZeDiKoBa19,KaBaDiJuKo21},
        LP- and MILP-solver based
        approaches~\cite{BuTuToKoMu18,Ehlers2017,TjXiTe19}, symbolic
        interval propagation~\cite{WaPeWhYaJa18}, abstraction and
        abstract-interpretation based
        techniques~\cite{ElGoKa20,GeMiDrTsChVe18,AsHaKrMo20},
        techniques for tackling recurrent
        networks~\cite{ZhShGuGuLeNa20,JaBaKa20} and binarized
        networks~\cite{AmWuBaKa21,NaKaRySaWa17}, and many others
        (e.g.,~\cite{DuJhSaTi18,LoMa17,RuHuKw18}); and these
        techniques have been applied to multiple ends, such as
        verifying adversarial robustness
        properties~\cite{GeMiDrTsChVe18,TjXiTe19,GoKaPaBa18,KuKaGoJuBaKo18,CaKaBaDi17,KaBaDiJuKo17Fvav},
        verifying hybrid systems with DNN
        controllers~\cite{DuChSa19,SuKhSh19}, verifying DNNs that
        serve as controllers for congestion control
        systems~\cite{KaBaKaSc19,AmScKa21,ElKaKaSc21}, and DNN
        simplification~\cite{GoFeMaBaKa20,LaKa21}.  As DNN
        verification engines continue to improve, so will the speed
        and scalability of our approach. Further, our line of work
        continues to demonstrate that DNN repair is an attractive
        application domain for verification.
	
	\section{Conclusion and Future Work}
	\label{sec:conclusion}
	
	Due to the recent surge in DNN popularity, it is becoming
	increasingly important to provide tools and methodologies for facilitating tasks
	that naturally arise as part of DNN usage --- such as modifying
	existing DNNs. Verification-based modification techniques offer
	significant advantages, and in this work, we have taken a step towards
	improving their applicability. Specifically, we were able to move beyond
	the single-layer change barrier that existed in prior work, and
	propose an approach that can simultaneously modify multiple layers of
	the DNN. Consequently, our approach can find modifications that are
	superior to those that would have been discovered by existing
	techniques.
	
	Moving forward, we plan to extend our approach along several
	axes. First, we intend to explore additional strategies for conducting
	the grid search, as the strategy in use has a significant effect on
	overall performance. Specifically, we intend to train a \emph{DNN
		controller} to manage the search strategy. Second, we observe that
	the grid search naturally lends itself to parallelization, and so we
	intend to explore parallelization techniques; and third, we intend to
	further demonstrate the usefulness of our technique by applying it to
	additional DNNs and case studies.
	
	\subsubsection{Acknowledgements.}
        This work was partially supported by the Israel Science
        Foundation (grant number 683/18) and the HUJI Federmann Cyber
        Security Center.
	
	{
		\newpage
		\bibliographystyle{abbrv}
		\bibliography{article}
	}
\end{document}